\documentclass{article}
\usepackage{ijcai26}

\usepackage{times}
\usepackage{soul}
\usepackage{url}
\usepackage[hidelinks]{hyperref}
\usepackage[utf8]{inputenc}
\usepackage[small]{caption}
\usepackage{graphicx}
\usepackage{amsmath}
\usepackage{cleveref}
\usepackage{amsthm}
\usepackage{booktabs}
\usepackage{algorithm}
\usepackage{algorithmic}
\usepackage[switch]{lineno}

\usepackage{subcaption} 

\usepackage[table]{xcolor}

\usepackage{multirow}
\usepackage{amsmath}
\usepackage{amssymb}

\definecolor{bestc}{HTML}{69b3a2} 
\definecolor{goodc}{HTML}{b4dcd3} 
\definecolor{badc}{HTML}{f8c8c8}
\definecolor{worstc}{HTML}{eeb4b4}
\definecolor{imagegreen}{HTML}{ADEBE6}
\definecolor{darker_green1}{HTML}{8CD1CD}
\definecolor{darker_green2}{HTML}{7BC2B8}
\definecolor{gray2}{HTML}{D9D9D9} 
\definecolor{gray1}{HTML}{F1F3F5} 
\definecolor{darkblue}{HTML}{00008B}

\title{Don’t Break the Boundary: Continual Unlearning for OOD Detection Based on Free Energy Repulsion}

\author{
Ningkang Peng$^1$\and
Kun Shao$^1$\and
Jingyang Mao$^1$\and
Linjing Qian$^1$\and
Xiaoqian Peng$^2$\and\\
Xichen Yang$^{1,*}$\and
Yanhui Gu$^{1}$\thanks{Corresponding authors.}
\\
\affiliations
$^1$School of Computer and Electronic Information, Nanjing Normal University, China\\
$^2$School of Artificial Intelligence and Information Technology, Nanjing University of Chinese Medicine, China\\
\emails
\{xichen\_yang, gu\}@njnu.edu.cn
}

\begin{document}

\maketitle

\begin{abstract}
Deploying trustworthy AI in open-world environments faces a dual challenge: the necessity for robust Out-of-Distribution (OOD) detection to ensure system safety, and the demand for flexible machine unlearning to satisfy privacy compliance and model rectification. However, this objective encounters a fundamental geometric contradiction: current OOD detectors rely on a static and compact data manifold, whereas traditional classification-oriented unlearning methods disrupt this delicate structure, leading to a catastrophic loss of the model's capability to discriminate anomalies while erasing target classes. To resolve this dilemma, we first define the problem of boundary-preserving class unlearning and propose a pivotal conceptual shift: in the context of OOD detection, effective unlearning is mathematically equivalent to transforming the target class into OOD samples. Based on this, we propose the TFER (Total Free Energy Repulsion) framework. Inspired by the free energy principle, TFER constructs a novel Push-Pull game mechanism: it anchors retained classes within a low-energy ID manifold through a pull mechanism, while actively expelling forgotten classes to high-energy OOD regions using a free energy repulsion force. 
This approach is implemented via parameter-efficient fine-tuning, circumventing the prohibitive cost of full retraining. Extensive experiments demonstrate that TFER achieves precise unlearning while maximally preserving the model's discriminative performance on remaining classes and external OOD data. More importantly, our study reveals that the unique Push-Pull equilibrium of TFER endows the model with inherent structural stability, allowing it to effectively resist catastrophic forgetting without complex additional constraints, thereby demonstrating exceptional potential in continual unlearning tasks.
\end{abstract}

\begin{figure}[ht]
    \centering
    \begin{subfigure}[t]{0.48\columnwidth} 
        \centering
        \includegraphics[width=\textwidth]{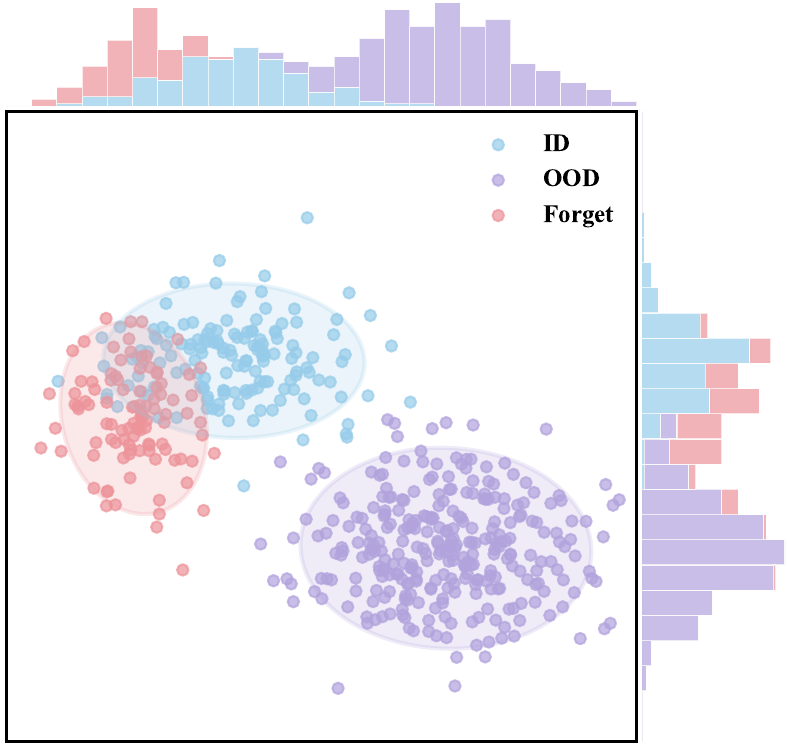} 
        \caption{Baseline}
        \label{fig:umap:baseline}
    \end{subfigure}
    \hfill 
    \begin{subfigure}[t]{0.48\columnwidth}
        \centering
        \includegraphics[width=\textwidth]{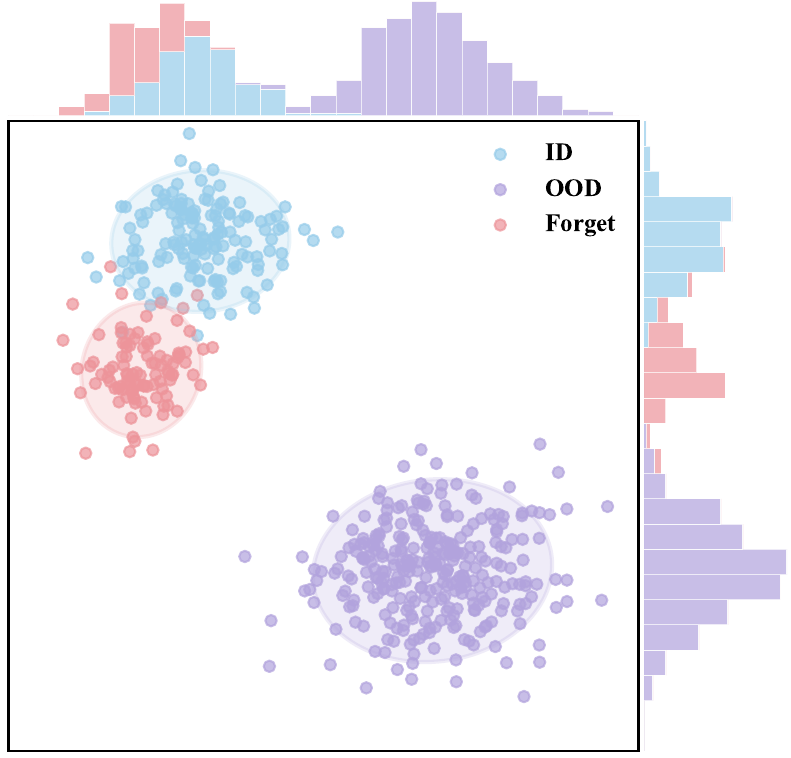} 
        \caption{TFER (Ours)}
        \label{fig:umap:tfer}
    \end{subfigure}
    
    \caption{
        \textbf{Visualizing Boundary-Preserving Unlearning: UMAP Projection of the Feature Manifold.}
        This figure compares the structural changes in the feature space after class unlearning, showcasing the Retained ID classes (Blue), External OOD samples (Purple), and the Target Forgotten class (Red).
    }
    \label{fig:umap}
\end{figure}

\section{Introduction}

In recent years, distance-based Out-of-Distribution (OOD) detection\cite{zhang2024conc} methods have emerged as the safety cornerstone for open-world perception tasks by constructing compact in-distribution (ID) data manifolds\cite{c:40,c:41,c:42}. However, existing OOD detection paradigms rely on a fragile static assumption: that the set of ID classes remains fixed post-deployment\cite{static-assumption,static-assumption2}. This premise stands in fundamental conflict with real-world dynamic unlearning needs arising from data privacy regulations (e.g., GDPR) and model correction requirements \cite{open-world-needs,MU}. This exposes a critical yet underexplored contradiction: traditional unlearning methods, typically designed for classification tasks, aim primarily to reduce classification accuracy on forgotten classes, yet they completely overlook the precise feature boundaries upon which OOD detection relies \cite{overlook-feature-boundaries,overlook-feature-boundaries2}. How to precisely excise specific classes from the feature manifold without compromising existing OOD detection capabilities has emerged as a core challenge demanding urgent solution.

As illustrated in Fig.\ref{fig:umap}, directly applying these conventional methods to OOD detection models induces catastrophic distortion of the ID manifold. Simple gradient ascent or noisy fine-tuning not only leads to manifold collapse but also forces forgotten and retained classes to overlap in the feature space. This blurs the decision boundary between ID and external OOD samples, ultimately causing the model to fail completely against unknown threats \cite{manifold-distorted,manifold-distorted2}.

Therefore, we argue that unlearning in the context of OOD detection is fundamentally a boundary reshaping problem. A successful solution must navigate a Trilemma \cite{trilemma}: simultaneously achieving (1)\textbf{ Unlearning Effectiveness}: thoroughly converting forgotten classes into OOD, (2) \textbf{Utility Preservation}: maintaining the compactness of retained ID classes, and (3) \textbf{OOD Generalization}: preserving detection capability against external anomalies. We propose a core thesis: in the OOD detection context, successful unlearning is equivalent to OOD transformation (Forget-as-OOD); thus, its effectiveness must be defined by OOD metrics rather than classification accuracy. To address this challenge, we introduce the TFER paradigm. This is a Push-Pull paradigm based on Low-Rank Adaptation (LoRA) \cite{lora}, designed to precisely isolate forgotten classes without disrupting the integrity of the ID manifold. Furthermore, we extend TFER to the scenario of continual unlearning. Unlike static single-task settings, sequential unlearning requests introduce the risk of 'Catastrophic Recall,' where parameter interference during the optimization of new targets can inadvertently reactivate previously unlearned information. To address this, we propose a Modular Orthogonal Strategy. Instead of relying on a single shared module, we introduce task-specific LoRA adapters constrained by gradient orthogonality. This design ensures that the optimization subspace of the current unlearning task remains geometrically isolated from historical ones. By enforcing orthogonality, TFER effectively prevents the unlearning operation of one concept from compromising the efficacy of previous unlearning efforts, establishing a robust paradigm for sustainable, lifelong machine unlearning.

Our main contributions are as follows: \begin{itemize} 
\item We formally define the problem of boundary-preserving unlearning in OOD detection for the first time and establish the Forget-as-OOD evaluation criterion. 
\item We propose the TFER loss, which leverages the principle of free energy repulsion to construct a universal anti-likelihood optimization objective, effectively resolving the manifold collapse issue caused by traditional methods. 
\item We design a LoRA-based Push-Pull framework that not only achieves SOTA performance in single-task unlearning but also demonstrates intrinsic stability and efficiency in continual unlearning scenarios .
\end{itemize}

\section{Related Work}
\label{sec:related_work}

\subsection{OOD Detection}
\label{sec:ood_detection}
OOD detection is key to ensuring the safe deployment of deep learning systems in open-world scenarios\cite{c:25,open-world-needs}. The field has developed various strategies, including methods based on confidence scores\cite{confidence-score2}, density estimation\cite{flowbased-density,density2,density3}, and energy models\cite{EBM-liu2020energy,EBM2}. In recent years, distance-based and prototype-based\cite{c:33} methods have emerged as a SOTA paradigm. The core idea of these methods is to learn and construct a compact ID feature manifold. OOD samples are subsequently identified by measuring their significant distance from all learned class prototypes or the manifold itself\cite{c:33,c:34,c:62}.
However, the efficacy of these detectors entirely relies on the integrity of this precisely learned ID manifold. This probabilistic boundary is highly sensitive and extremely susceptible to distortion from naive model updates. While these OOD detectors are powerful, their static nature and their boundary sensitivity present a high-risk and unresolved challenge when facing the need for selective data removal or model rectification.

\subsection{Machine Unlearning}
\label{sec:machine_unlearning}
Machine Unlearning (MU) \cite{largeUNLEARNING} aims to remove the influence of specific data from a trained model, driven by needs for data privacy \cite{MU}, bias mitigation, or error correction. Due to the high cost of full retraining, research mainly focuses on approximate unlearning methods.
Existing approximate unlearning methods can be broadly categorized into: architecture-based strategies (e.g., sharded training \cite{MU-SISA}), methods based on regularization or knowledge distillation (aimed at preserving remaining knowledge), and model finetuning-based strategies. The latter simulate the unlearning process through specific loss functions or gradient operations, such as using gradient ascent to reverse the learning process \cite{ga,ga2} or corrupting knowledge with random labels. These finetuning-based methods are commonly used as effective baselines for unlearning tasks due to their ease of integration into deep network training pipelines.
However, all existing unlearning techniques suffer from a fundamental limitation: they are classification-centric. Their optimization objective is to degrade the classification accuracy for the forgotten data \cite{bad-optimization-objective,bad-optimization-objective2,trilemma}, and they are unable to actively push the forgotten data off the ID manifold. This fundamental misalignment in objectives means that all traditional methods fail to solve the problem of boundary-preserving unlearning in OOD detection, as they are insufficient to maintain the geometric integrity and probabilistic density of the feature manifold required by OOD detectors.

\begin{figure*}[!ht]
    \centering \includegraphics[width=1\textwidth]{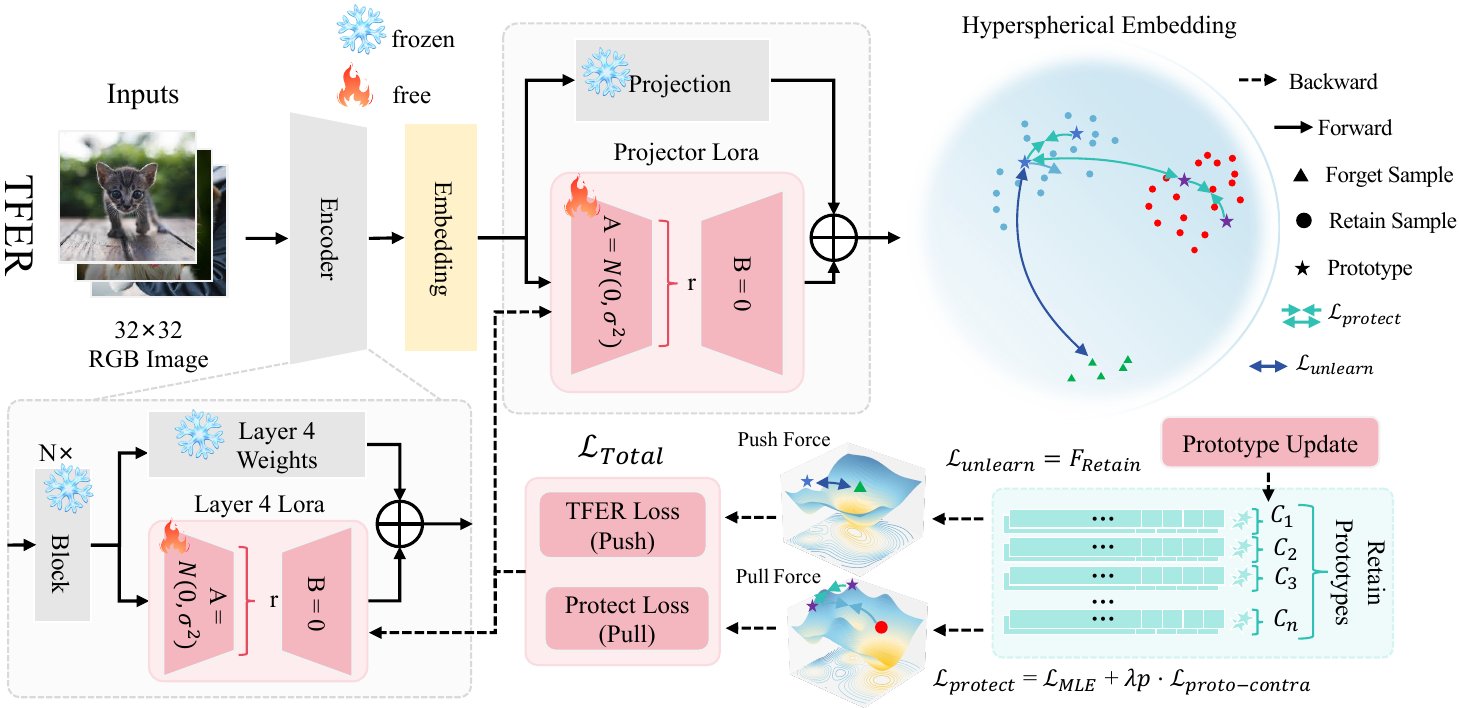} 
        \caption{
        The framework decomposes the class unlearning problem into an adversarial process of \textbf{Push Force and Pull Force} in a high-dimensional hyperspherical embedding space, implemented via parameter-efficient LoRA modules. The Push Force ($\mathcal{L}_{\text{unlearn}}$), represented by the TFER loss, pushes \textbf{Forget Samples} away from all Retain Prototypes. The Pull Force ($\mathcal{L}_{\text{protect}}$), the protection loss, pulls Retain Samples towards their corresponding prototypes and optimizes the prototype structure via the prototype contrastive loss ($\mathcal{L}_{\text{proto-contra}}$) to ensure clear separation of retain classes.
    }
    \label{fig:framework}
\end{figure*}

\section{Methodology}
\label{sec:method}

As illustrated in Fig.\ref{fig:framework}, we propose the TFER framework, a geometric unlearning paradigm specifically designed to resolve the boundary reshaping trilemma. Unlike naive unlearning approaches that indiscriminately disrupt feature filters, TFER orchestrates a Push-Pull dynamic: a \textbf{Push Force} that expels forgetting samples into high-energy regions, and a \textbf{Pull Force} that anchors the retained manifold to preserve utility. The core philosophy is \textbf{Forget-as-OOD}: transforming the forgetting class $\mathcal{Y}_{\text{Forget}}$ into a high-energy state that is statistically indistinguishable from natural outliers.

\subsection{Preliminaries: Prototype-based OOD Detection}
\label{sec:method:preliminaries}
We instantiate TFER on a hyperspherical OOD detection framework. For each class $j \in \mathcal{Y}_{\text{ID}}$, the feature distribution is modeled as a mixture of $K$ von Mises-Fisher (vMF) distributions on a unit $(D-1)$-manifold. A vMF distribution $vMF(\mathbf{z}; \mathbf{\mu}, \kappa)$ for a feature $\mathbf{z} \in \mathbb{S}^{D-1}$ is defined as:
\begin{equation}
    p(\mathbf{z} | \mathbf{\mu}, \kappa) = C_D(\kappa) \exp(\kappa \mathbf{\mu}^\top \mathbf{z}),
    \label{eq:vmf_pdf}
\end{equation}
where $\mathbf{\mu}$ is the mean direction, $\kappa \geq 0$ is the concentration parameter, and $C_D(\kappa)$ is a normalization constant.

In our framework, each class $j$ is represented by a set of prototypes $\{\mathbf{p}_k^j\}_{k=1}^K$ acting as mean directions. The unnormalized log-likelihood (logit) for a feature $\mathbf{z}$ is then defined via the mixture:
\begin{equation}
    L_j(\mathbf{z}) = \log \sum_{k=1}^K w_{\mathbf{z},k}^j \exp(\kappa (\mathbf{p}_k^j)^\top \mathbf{z}),
    \label{eq:vmf_logit}
\end{equation}
where $w_{\mathbf{z},k}^j$ is the posterior assignment weight of $\mathbf{z}$ to the $k$-th prototype. In this hyperspherical space, OOD detection is performed by thresholding the negative free energy or maximum similarity across all ID manifolds.

\subsection{The Push Mechanism: Energy Barrier Construction}
\label{sec:method:push}
Following the Forget-as-OOD strategy, to excise $\mathcal{Y}_{\text{Forget}}$, we adopt an Energy-Based Model (EBM) perspective. Based on the Gibbs distribution $p(\mathbf{z}) \propto \exp(-E(\mathbf{z}))$, a sample is recognized as ID only if it resides in a low-energy valley. 

Our goal is to erect an energy barrier for the forgetting class. We define the Total Free Energy of a sample relative to the retained system as the Log-Sum-Exp of its retained logits:
\begin{equation}
    E(\mathbf{z}; \mathcal{Y}_{\text{Retain}}) = -\log \sum_{j \in \mathcal{Y}_{\text{Retain}}} \exp(L_j(\mathbf{z})).
\end{equation}
By minimizing the negative free energy, we derive the TFER Push Loss:
\begin{equation}
    \mathcal{L}_{\text{unlearn}}(\mathbf{z}_u) = \log \sum_{j \in \mathcal{Y}_{\text{Retain}}} \exp(L_j(\mathbf{z}_u)).
    \label{eq:tfer_loss1}
\end{equation}
Unlike classification-based unlearning which merely reassigns labels, Eq.\eqref{eq:tfer_loss1} forces $\mathbf{z}_u$ to drift away from \textit{all} retained class manifolds simultaneously. This ensures that $\mathbf{z}_u$ is pushed into the rejection region typically occupied by external OOD samples.

\subsection{Theoretical Analysis: Gradient Stability via Convex Geometry}
\label{sec:method:gradient_analysis}

A prevalent issue in unlearning tasks is the instability induced by \textit{gradient conflict}, which is particularly evident in naive Gradient Ascent (GradAsc) approaches that attempt to directly maximize the loss of the forgetting class. Here, we provide a geometric analysis to demonstrate why TFER achieves superior stability.
Consider the gradient of $\mathcal{L}_{\text{unlearn}}$ with respect to the feature $\mathbf{z}_u$:
\begin{equation}
    \nabla_{\mathbf{z}_u} \mathcal{L}_{\text{unlearn}} = \sum_{j \in \mathcal{Y}_{\text{Retain}}} \underbrace{\left[ \frac{e^{L_j(\mathbf{z}_u)}}{\sum_{i \in \mathcal{Y}_{\text{Retain}}} e^{L_i(\mathbf{z}_u)}} \right]}_{\alpha_j(\mathbf{z}_u)} \cdot \nabla_{\mathbf{z}_u} L_j(\mathbf{z}_u),
    \label{eq:tfer_grad_final1}
\end{equation}
where $\alpha_j(\mathbf{z}_u) \in (0,1)$ represents the importance weight of retain class $j$, satisfying $\sum \alpha_j = 1$. This formulation reveals two critical geometric properties that guarantee stability:

\paragraph{Boundedness via Convex Hull Constraint.}
Unlike GradAsc, where the gradient norm can grow unboundedly as the model yields overconfident erroneous predictions, Eq.\eqref{eq:tfer_grad_final1} indicates that the update direction of TFER is strictly confined within the Convex Hull of the gradients of the retained classes:
\begin{equation}
    \nabla_{\mathbf{z}_u} \mathcal{L}_{\text{unlearn}} \in \text{Conv}(\{ \nabla L_j \}).
\end{equation}
This acts as an implicit gradient clipping mechanism. Even if the repulsion force from a specific class is extremely strong, the aggregate gradient is smoothed via convex combination, thereby preventing the gradient explosion issue that frequently disrupts feature manifolds in traditional unlearning.

\paragraph{Orthogonality and Entropy Maximization.}
The coefficients $\alpha_j(\mathbf{z}_u)$ govern the directionality. When $\mathbf{z}_u$ is deeply situated within the manifold of a specific retained class $k$, the gradient is dominated by $\nabla L_k$. This implies that the force aligns with the normal direction, specifically pushing the sample away from the nearest boundary. However, as $\mathbf{z}_u$ is pushed away and becomes equidistant to multiple retained classes (the ideal OOD state), the weights $\alpha_j$ tend toward a uniform distribution. At this stage, the resultant force becomes the average of all repulsion forces, effectively propelling the sample into the void region orthogonal to the entire ID manifold, rather than oscillating between specific classes.
By directing the forgotten samples toward the null space of the ID manifold, TFER effectively mimics the behavior of natural anomalies, thereby fulfilling the Forget-as-OOD objective without encroaching upon the decision boundaries of existing OOD detectors.

\subsection{The Pull Mechanism: Manifold Anchoring}
To prevent \textit{Manifold Drift} of the retained classes, we introduce a prototype-based manifold anchoring mechanism. Assuming each retained class $j$ is represented by a prototype vector $\mathbf{p}_j$, we define the Pull loss $\mathcal{L}_{\text{protect}}$ as the negative log-likelihood:
\begin{equation}
\scalebox{0.9}{%
    $\displaystyle \mathcal{L}_{\text{protect}}(\mathcal{D}_{\text{Retain}}) = \mathbb{E}_{(\mathbf{x}, y) \sim \mathcal{D}_{\text{Retain}}} \left[ -\log \frac{\exp(\mathbf{z}^\top \mathbf{p}_y / \tau)}{\sum_{k \in \mathcal{Y}_{\text{Retain}}} \exp(\mathbf{z}^\top \mathbf{p}_k / \tau)} \right],$ 
}
\end{equation}
where $\tau$ is the temperature coefficient. This term anchors the feature manifold within the $\epsilon$-neighborhood of the pre-trained topology $\mathcal{T}_0$ by maximizing the cosine similarity between retained samples and their corresponding prototypes.

\subsection{Optimization Strategy: Rank-Constrained Adaptation}
To efficiently solve the aforementioned Push-Pull objective within the parameter space $\Theta$, we model the unlearning process as a Rank-Constrained Optimization problem.

We decompose the parameter update as $\Theta = \Theta_0 + \Delta \Theta$. To preserve the general feature extraction capabilities of $\Theta_0$, we freeze $\Theta_0$ and enforce a low-rank constraint on $\Delta \Theta$:
\begin{equation}
    \Delta \Theta = B A, \quad \text{where } B \in \mathbb{R}^{d \times r}, A \in \mathbb{R}^{r \times k}, r \ll d.
    \label{eq:lora_def}
\end{equation}
The final optimization objective is to find the optimal low-rank matrices $A^*, B^*$:
\begin{equation}
    \min_{A, B} \quad \mathcal{J} = \lambda_f \mathcal{L}_{\text{unlearn}}(\mathcal{D}_{\text{Forget}}; \Theta) + \mathcal{L}_{\text{protect}}(\mathcal{D}_{\text{Retain}}; \Theta).
    \label{eq:final_objective}
\end{equation}
The low-rank decomposition not only reduces computational cost but, more importantly, restricts the Frobenius norm $\|\Delta \Theta\|_F$ of the parameter updates. This introduces implicit regularization into the solution space, ensuring that the model does not suffer from catastrophic forgetting while excising the forgotten knowledge. In our exploration of continual unlearning, we introduce a constraint term $\mathcal{L}_{orth}$ that enforces the update vector of the new task $\Delta A = A_t - A_{t-1}^{ref}$ to remain geometrically orthogonal to the reference matrix $A_{t-1}^{ref}$ which stores historical knowledge. The orthogonal loss function is defined as follows:

\begin{equation}
    \mathcal{L}_{orth} = \| (A_t - A_{t-1}^{ref})^{\top} A_{t-1}^{ref} \|_F^2
    \label{eq:orthogonal_loss}
\end{equation}

\begin{table*}[ht]
\centering

\resizebox{\textwidth}{!}{%
\begin{tabular}{llccccccccccccccc}
\toprule
\multirow{3}{*}{\textbf{Forgetting}} & \multirow{3}{*}{\textbf{Method}} & \multicolumn{2}{c}{\textbf{Unlearning Efficacy}} & \multicolumn{6}{c}{\textbf{Utility Preservation}} & \multicolumn{2}{c}{\textbf{AVG (OOD)}} \\

 \multirow{3}{*}{\hspace*{10pt}\textbf{Class}}& & & & Retain- & \multicolumn{1}{c}{SVHN} & \multicolumn{1}{c}{Places365} & \multicolumn{1}{c}{LSUN} & \multicolumn{1}{c}{iSUN} & \multicolumn{1}{c}{Textures} & \multirow{2}{*}{AUC} & \multirow{2}{*}{FPR} \\

 & & AUC $\uparrow$ & FPR $\downarrow$ & Acc $\uparrow$ & AUC/FPR & AUC/FPR & AUC/ FPR & AUC/FPR & AUC/FPR & & \\
\midrule

 & Original & 87.1 & 82.3 & 74.5 & 99.3 / 2.6 & 80.9 / 68.7 & 95.0 / 21.9 & 81.4 / 72.9 & 90.1 / 39.0 & 89.3 & 41.0 \\
 & Retrain & 53.2 & 96.8 & 48.1 & 92.1 / 32.1 & 64.2 / 88.1 & 60.3 / 84.2 & 64.2 / 92.0 & 88.1 / 41.5 & 73.8 & 67.6 \\
 & RL-FT & 90.6 & 39.4 & 73.6 & 98.3 / 2.6 & 80.3 / 67.8 & 96.0 / 18.2 & 80.1 / 73.1 & 89.7 / 42.1 & 88.9 & 40.8 \\
 & GradAsc & 74.7 & 71.8 & 72.8 & 98.4 / 7.0 & 78.2 / 72.3 & 93.4 / 30.5 & 80.9 / 73.3 & 89.3 / 44.1 & 88.0 & 45.4 \\
 \rowcolor{gray1}\multirow{-5}{*}{\hspace*{20pt}\textbf{5}} & \textbf{TFER (Ours)} & \textcolor{darkblue}{\textbf{91.7}} & \textcolor{darkblue}{\textbf{37.9}} & \textcolor{darkblue}{\textbf{74.8}} & \textbf{99.3} / \textbf{2.4} & \textbf{80.3} / \textbf{67.8} & \textbf{96.2} / \textbf{18.2} & \textbf{81.0} / \textbf{75.3} & \textbf{90.7} / \textbf{39.7} & \textcolor{darkblue}{\textbf{89.5}} & \textcolor{darkblue}{\textbf{40.7}} \\
\midrule

 & Original & 87.1 & 83.5 & 74.5 & 99.3 / 2.6 & 80.9 / 68.7 & 95.0 / 21.9 & 81.4 / 72.9 & 90.1 / 39.0 & 89.4 & 41.0 \\
 & Retrain & 55.0 & 94.9 & 40.1 & 89.9 / 36.1 & 60.2 / 92.2 & 57.4 / 87.3 & 61.6 / 95.0 & 87.1 / 43.9 & 71.2 & 70.9 \\
 & RL-FT & 90.2 & 49.2 & 75.4 & 97.6 / 4.4 & 78.7 / 72.3 & 93.2 / 32.6 & 83.6 / 69.7 & 89.5 / 45.2 & 88.5 & 44.8 \\
 & GradAsc & 73.5 & 81.8 & 73.7 & 96.4 / 17.9 & 78.3 / 75.4 & 88.9 / 53.7 & 83.5 / 69.1 & 87.3 / 55.6 & 86.9 & 54.3 \\
\rowcolor{gray1}  \multirow{-5}{*}{\hspace*{20pt}\textbf{10}} & \textbf{TFER (Ours)} & \textcolor{darkblue}{\textbf{91.3}} & \textcolor{darkblue}{\textbf{41.2}} & \textcolor{darkblue}{\textbf{74.2}} & \textbf{99.2} / \textbf{2.4} & \textbf{80.1} / \textbf{68.6} & \textbf{94.9} / \textbf{22.0} & \textbf{84.6} / \textbf{67.8} & \textbf{90.0} / \textbf{40.9} &\textcolor{darkblue}{ \textbf{89.8}} & \textcolor{darkblue}{\textbf{40.3} }\\
\midrule

 & Original & 87.5 & 84.1 & 75.6 & 99.3 / 2.6 & 80.9 / 68.7 & 95.0 / 21.9 & 81.4 / 72.9 & 90.1 / 39.0 & 89.3 & 41.0 \\
 & Retrain & 56.1 & 93.5 & 37.6 & 88.7 / 38.8 & 57.9 / 93.3 & 55.1 / 89.3 & 59.2 / 96.1 & 86.8 / 45.2 & 69.5 & 72.5 \\
 & RL-FT & 87.7 & 49.2 & 75.4 & 96.4 / 4.4 & 78.3 / 75.5 & 93.2 / 32.7 & 83.5 / 69.7 & 90.2 / 45.3 & 88.3 & 45.5 \\
 & GradAsc & 73.1 & 81.8 & 73.7 & 96.4 / 17.9 & 78.3 / 75.5 & 89.0 / 53.7 & 83.3 / 69.1 & 87.3 / 55.6 & 86.9 & 54.4 \\
\rowcolor{gray1} \multirow{-5}{*}{\hspace*{20pt}\textbf{15}}  & \textbf{TFER (Ours)} & \textcolor{darkblue}{\textbf{90.9}} & \textcolor{darkblue}{\textbf{43.2}} & \textcolor{darkblue}{\textbf{73.9}} & \textbf{99.2} / \textbf{2.4} & \textbf{80.0} / \textbf{68.6} & \textbf{94.7} / \textbf{22.1} & \textbf{83.6} / \textbf{69.1} & \textbf{91.0} / \textbf{39.9} & \textcolor{darkblue}{\textbf{89.7}} & \textcolor{darkblue}{\textbf{40.5}} \\
\midrule

 & Original & 84.9 & 85.0 & 73.9 & 99.3 / 2.6 & 80.9 / 68.7 & 95.0 / 21.9 & 81.4 / 72.9 & 90.1 / 39.0 & 89.3 & 41.0 \\
 & Retrain & 57.2 & 91.8 & 30.8 & 86.3 / 40.4 & 56.3 / 94.1 & 53.7 / 90.5 & 57.8 / 97.1 & 84.2 / 47.8 & 67.7 & 74.0 \\
 & RL-FT & 89.8 & 45.3 & 76.1 & 93.1 / 4.0 & 79.9 / 70.4 & 93.8 / 29.1 & 87.3 / 55.4 & 88.8 / 46.6 & 88.6 & 41.1 \\
 & GradAsc & 72.5 & 71.1 & 71.9 & 95.5 / 22.7 & 79.6 / 73.1 & 87.4 / 57.0 & 88.3 / 55.1 & 85.7 / 62.6 & 87.3 & 54.1 \\
\rowcolor{gray1} \multirow{-5}{*}{\hspace*{20pt}\textbf{20}}  & \textbf{TFER (Ours)} & \textcolor{darkblue}{\textbf{90.7}} & \textcolor{darkblue}{\textbf{44.1}} & \textcolor{darkblue}{\textbf{76.1}} & \textbf{99.3} / \textbf{2.4} & \textbf{81.5} / \textbf{67.7} & \textbf{95.8} / \textbf{19.1} & \textbf{87.3} / \textbf{55.4} & \textbf{91.1} / \textbf{39.4} & \textcolor{darkblue}{\textbf{91.0}} & \textcolor{darkblue}{\textbf{36.8} }\\
\bottomrule
\end{tabular}
}
\caption{Main results comparing TFER with all baselines on CIFAR-100 under the 25 epoch setting. FPR denotes the False Positive Rate at 95\% Recall.}
\label{tab:main_results25}
\end{table*}

\section{Experiments}
\label{sec:experiments}

\subsection{Experimental Setup}
\label{sec:setup_revised}

We detail the experimental configuration used in this study, covering datasets, evaluation metrics, and baselines.

\subsubsection{Datasets.} We use CIFAR-100 as the primary ID dataset, randomly select $K$ classes as the target forgetting set $\mathcal{Y}_{\mathrm{Forget}}$, with the remaining classes forming the retained set $\mathcal{Y}_{\mathrm{Retain}}$. To evaluate the OOD detection capability after unlearning, we utilize standard external OOD test sets, including SVHN, LSUN, Textures, Places365, and iSUN.


\subsubsection{Evaluation Metrics.}
We evaluate performance across three dimensions:
(1) \textbf{Unlearning Completeness}: Following the Forget-as-OOD paradigm, we measure the rejection of $\mathcal{Y}_{Forget}$ using AUROC ($\uparrow$) and FPR95 ($\downarrow$) to quantify feature space divergence.
(2) \textbf{Utility \& Generalization}: To ensure discriminative power is preserved, we report classification accuracy on $\mathcal{Y}_{\text{Retain}}$ (Retain-Acc $\uparrow$) and average AUROC ($\uparrow$) on external OOD datasets.
(3) \textbf{Efficiency}: We report trainable parameters and training time to demonstrate the lightweight nature of our approach.

\subsubsection{Comparison Methods}
\label{sec:methods_revised}


We compare TFER with four representative baselines:
(1) \textbf{Original}: The pre-trained model without unlearning, serving as the upper bound for utility preservation.
(2) \textbf{Cost-Constrained Retrain}: The model trained from scratch solely on $\mathcal{Y}_{\mathrm{Retain}}$. To ensure a fair comparison of computational budgets, this baseline is trained for the same number of epochs as the unlearning methods. It represents the practical utility recovery achievable through retraining within a fixed time window.
(3) \textbf{GradAsc} \cite{ga}: A general unlearning method that maximizes the loss on $\mathcal{Y}_{\mathrm{Forget}}$ via gradient ascent.
(4) \textbf{Random Label FT}(RL-FT) \cite{MU}: Fine-tunes the model by assigning random labels to $\mathcal{Y}_{\mathrm{Forget}}$ samples.

\subsubsection{Implementation Details}
\label{sec:details_revised}

Following prior work, we employ ResNet-34 as the backbone for CIFAR-100, followed by an MLP projector to map features onto a hypersphere. During the unlearning phase, we freeze the backbone and apply LoRA to layer4 and the projector. We use an SGD optimizer with a constant learning rate of $3 \times 10^{-4}$ for 50 epochs. The primary OOD scoring function is the Mahalanobis distance computed on the penultimate layer features.

\subsection{Main Results and Analysis}
\label{sec:main_results}

\begin{table*}[ht]
\centering
\resizebox{\textwidth}{!}{%
\begin{tabular}{llccccccccccccccc}
\toprule
\multirow{3}{*}{\textbf{Forgetting}} & \multirow{3}{*}{\textbf{Method}} & \multicolumn{2}{c}{\textbf{Unlearning Efficacy}} & \multicolumn{6}{c}{\textbf{Utility Preservation}} & \multicolumn{2}{c}{\textbf{AVG (OOD)}} \\

 \multirow{3}{*}{\hspace*{10pt}\textbf{Class}}& & & & Retain- & \multicolumn{1}{c}{SVHN} & \multicolumn{1}{c}{Places365} & \multicolumn{1}{c}{LSUN} & \multicolumn{1}{c}{iSUN} & \multicolumn{1}{c}{Textures} & \multirow{2}{*}{AUC} & \multirow{2}{*}{FPR} \\

 & & AUC $\uparrow$ & FPR $\downarrow$ & Acc $\uparrow$ & AUC/FPR & AUC/FPR & AUC/ FPR & AUC/FPR & AUC/FPR & & \\
\midrule

 & Original & 87.1 & 82.3 & 74.5 & 99.3 / 2.6 & 80.9 / 68.7 & 95.0 / 21.9 & 81.4 / 72.9 & 90.1 / 39.0 & 89.3 & 41.0 \\
 & Retrain & 53.2 & 96.8 & 48.1 & 92.1 / 32.1 & 64.2 / 88.1 & 60.3 / 84.2 & 64.2 / 92.0 & 88.1 / 41.5 & 73.8 & 67.6 \\
 & RL-FT & 90.2 & 47.5 & 75.6 & 98.1 / 4.8 & 79.8 / 71.8 & 94.6 / 25.8 & 84.5 / 66.5 & 89.6 / 45.5 & 89.3 & 42.9 \\
 & GradAsc & 75.4 & 71.5 & 72.8 & 98.1 / 7.9 & 78.5 / 72.5 & 93.5 / 31.8 & 80.4 / 74.1 & 89.0 / 44.8 & 88.0 & 46.5 \\
 \rowcolor{gray1}\multirow{-5}{*}{\hspace*{20pt}\textbf{5}} & \textbf{TFER (Ours)} & \textcolor{darkblue}{\textbf{92.1}} & \textcolor{darkblue}{\textbf{38.5}} & \textcolor{darkblue}{\textbf{75.4}} & \textbf{99.4} / \textbf{2.3} & \textbf{79.5} / \textbf{67.5} & \textbf{96.5} / \textbf{18.0} & \textbf{81.5} / \textbf{75.0} & \textbf{90.9} / \textbf{39.5} & \textcolor{darkblue}{\textbf{89.6}} & \textcolor{darkblue}{\textbf{40.5}} \\
\midrule

 & Original & 87.1 & 83.5 & 74.5 & 99.4 / 2.6 & 80.9 / 68.7 & 95.0 / 21.9 & 81.4 / 72.9 & 90.1 / 39.0 & 89.4 & 41.0 \\
 & Retrain & 55.0 & 94.9 & 40.1 & 89.9 / 36.1 & 60.2 / 92.2 & 57.4 / 87.3 & 61.6 / 95.0 & 87.1 / 43.9 & 71.2 & 70.9 \\
 & RL-FT & 89.5 & 46.8 & 73.2 & 97.4 / 4.6 & 78.5 / 71.8 & 93.8 / 28.1 & 83.5 / 68.2 & 89.9 / 41.0 & 88.6 & 42.7 \\
 & GradAsc & 77.2 & 76.5 & 71.5 & 98.5 / 5.1 & 78.4 / 72.6 & 94.8 / 24.5 & 82.5 / 71.5 & 87.6 / 49.2 & 88.5 & 44.8 \\
\rowcolor{gray1}  \multirow{-5}{*}{\hspace*{20pt}\textbf{10}} & \textbf{TFER (Ours)} & \textcolor{darkblue}{\textbf{92.8}} & \textcolor{darkblue}{\textbf{39.2}} & \textcolor{darkblue}{\textbf{74.9}} & \textbf{98.9} / \textbf{3.1} & \textbf{80.3} / \textbf{68.5} & \textbf{95.5} / \textbf{21.8} & \textbf{84.0} / \textbf{67.5} & \textbf{91.3} / \textbf{38.8} &\textcolor{darkblue}{ \textbf{90.0}} & \textcolor{darkblue}{\textbf{40.0} }\\
\midrule

 & Original & 87.5 & 84.1 & 75.6 & 99.3 / 2.6 & 80.9 / 68.7 & 95.0 / 21.9 & 81.4 / 72.9 & 90.1 / 39.0 & 89.3 & 41.0 \\
 & Retrain & 56.1 & 93.5 & 37.6 & 88.7 / 38.8 & 57.9 / 93.3 & 55.1 / 89.3 & 59.2 / 96.1 & 86.8 / 45.2 & 69.5 & 72.5 \\
 & RL-FT & 90.8 & 47.8 & 75.4 & 98.6 / 4.8 & 79.1 / 71.9 & 94.5 / 25.8 & 84.5 / 66.6 & 89.6 / 45.6 & 89.3 & 42.9 \\
 & GradAsc & 75.2 & 79.2 & 73.1 & 96.1 / 18.5 & 78.5 / 76.2 & 90.2 / 51.5 & 84.2 / 67.5 & 87.8 / 54.6 & 87.4 & 53.9 \\
\rowcolor{gray1} \multirow{-5}{*}{\hspace*{20pt}\textbf{15}}  & \textbf{TFER (Ours)} & \textcolor{darkblue}{\textbf{91.0}} & \textcolor{darkblue}{\textbf{42.1}} & \textcolor{darkblue}{\textbf{75.3}} & \textbf{99.3} / \textbf{2.6} & \textbf{80.2} / \textbf{68.5} & \textbf{95.1} / \textbf{22.0} & \textbf{85.1} / \textbf{66.0} & \textbf{91.2} / \textbf{39.6} & \textcolor{darkblue}{\textbf{89.9}} & \textcolor{darkblue}{\textbf{41.2}} \\
\midrule

 & Original & 84.9 & 85.0 & 73.9 & 99.3 / 2.6 & 80.9  68.7 & 95.0 / 21.9 & 81.4 / 72.9 & 90.1 / 39.0 & 89.3 & 41.0 \\
 & Retrain & 57.2 & 91.8 & 30.8 & 86.3 / 40.4 & 56.3 / 94.1 & 53.7 / 90.5 & 57.8 / 97.1 & 84.2 / 47.8 & 67.7 & 74.0 \\
 & RL-FT & 90.8 & 45.8 & 75.8 & 98.8 / 4.3 & 80.5 / 68.9 & 94.4 / 25.7 & 88.8 / 49.2 & 89.2 / 47.5 & 90.3 & 39.1 \\
 & GradAsc & 79.5 & 67.1 & 73.8 & 95.2 / 23.0 & 78.5 / 75.0 & 84.1 / 64.9 & 88.2 / 53.6 & 85.2 / 65.6 & 86.2 & 56.8 \\
 \rowcolor{gray1} \multirow{-5}{*}{\hspace*{20pt}\textbf{20}}  & \textbf{TFER (Ours)} & \textcolor{darkblue}{\textbf{91.9}} & \textcolor{darkblue}{\textbf{41.5}} & \textcolor{darkblue}{\textbf{76.5}} & \textbf{99.4} / \textbf{2.3} & \textbf{81.6} / \textbf{67.5} & \textbf{96.0} / \textbf{18.9} & \textbf{88.5} / \textbf{48.5} & \textbf{89.9} / \textbf{44.5} & \textcolor{darkblue}{\textbf{91.1}} & \textcolor{darkblue}{\textbf{36.3} }\\
\bottomrule
\end{tabular}
}
\caption{Main results comparing TFER with all baselines on CIFAR-100 under the 50 epoch setting. FPR denotes the False Positive Rate at 95\% Recall. All results are percentages (\%).}
\label{tab:main_results50}
\end{table*}

\subsubsection{Overall Performance and Mechanism Analysis}
Tab.\ref{tab:main_results25} and Tab.\ref{tab:main_results50} present a comprehensive comparison on CIFAR-100. TFER consistently achieves a superior trade-off among unlearning efficacy, utility preservation, and robustness.

\paragraph{Unlearning Efficacy: Explicit Boundary Reshaping.}
TFER demonstrates a structural advantage in converting $\mathcal{Y}_{\mathrm{Forget}}$ into detectable OOD samples, significantly outperforming approximate baselines. As evidenced in Tab.\ref{tab:main_results50}, methods like RandLabel FT yield high FPR95 scores, suggesting that label noise merely introduces local decision ambiguity rather than creating separation. In contrast, TFER achieves the lowest FPR95, confirming that our $\mathcal{L}_{\mathrm{unlearn}}$ objective actively constructs a high-energy barrier around forgetting classes. This effectively pushes them away from the in-distribution manifold into low-density regions, achieving what gradient ascent fails to do due to optimization conflicts.

\paragraph{Utility Preservation: Manifold Anchoring via LoRA.} A critical observation is the catastrophic collapse of the cost-constrained Retrain baseline in complex settings, where accuracy drops to 32.50\%. This failure highlights that discarding the inductive bias of the pre-trained model makes reconstructing the feature manifold from random initialization intractable under limited training iterations. Conversely, TFER maintains a Retain Accuracy of 76.1\%, comparable to the original model. This near-lossless retention is attributed to our Manifold Anchoring mechanism: by freezing the pre-trained backbone, TFER preserves the integrity of the underlying representations while selectively excising target class information. This ensures the optimization trajectory corrects the decision boundary without the manifold collapse typically observed when retraining from scratch under strict resource constraints.

\paragraph{Scalability and Structural Stability.}
We stress-test structural stability by varying the forgetting scope (5 to 20 classes) and training duration. While the Retrain baseline exhibits high sensitivity where it performs adequately on 5 classes but collapses on 20, TFER maintains a remarkably flat performance trajectory. Even under high-forgetting stress ,such as 20 classes, the decoupling of unlearning dynamics from general feature extraction ensures stability. Furthermore, TFER achieves convergence at 25 epochs and further optimizes boundary distinction by 50 epochs, FPR95 drops from 44.1\% to 41.5\%, validating its scalability as an efficient, robust unlearning framework.

\paragraph{Robustness of OOD Generalization.}
Finally, external OOD metrics (AVG. OOD) reveal the integrity of the learned features. TFER consistently outperforms Retrain, indicating that Retrain overfits the reduced subset, yielding brittle decision boundaries, whereas TFER preserves the semantic density of the original pre-trained model. The unlearning process in TFER is strictly constrained within the known space, thereby maintaining the model's capability to detect unknown regions.



\subsection{Efficiency Analysis}
Compared to full retraining, TFER significantly curtails computational overhead and accelerates the unlearning process. As detailed in Tab.~\ref{tab:efficiency}, TFER reduces trainable parameters by 98.9\% relative to the original ResNet-34 backbone. The unlearning procedure converges in approximately 9 minutes, achieving a 20$\times$ speedup over the retraining baseline.

This efficiency gain is primarily attributed to TFER’s parameter-efficient adaptation strategy. While retraining necessitates back-propagation through all 21.6M parameters, incurring prohibitive memory and time costs, TFER employs low-rank subspace optimization by freezing the backbone and updating only the lightweight LoRA modules. From a geometric perspective, retraining attempts to reconstruct the entire feature manifold from a random initialization. In contrast, TFER treats unlearning as a localized boundary correction. By leveraging the pre-trained feature extraction priors, the Push-Pull dynamics efficiently reshape the decision boundaries within a constrained low-dimensional subspace, bypassing the intensive feature-relearning phase required by traditional methods.


\subsection{Ablation Studies}
\label{sec:ablation}

We conduct further ablation studies to validate the design choices of the TFER framework.

\subsubsection{Ablation on loss}

We validate the TFER Push-Pull paradigm by ablating its core loss terms. As shown in Tab.\ref{tab:ablation_2}, removing $\mathcal{L}_{\mathrm{unlearn}}$ causes the Forget FPR95 to surge to 84.50, as the model lacks the necessary gradient signal to expel target samples from their low-energy potential wells. Without this explicit repulsion, forgetting samples remain statistically indistinguishable from the ID manifold.

Crucially, omitting $\mathcal{L}_{\mathrm{protect}}$ also degrades unlearning efficacy, supporting our Manifold Anchor theory. While $\mathcal{L}_{\mathrm{unlearn}}$ provides the repulsive force, $\mathcal{L}_{\mathrm{protect}}$ establishes a rigid reference frame by preventing the retained manifold from drifting due to gradient oscillations.
In summary, the synergy between $\mathcal{L}_{\mathrm{unlearn}}$ and $\mathcal{L}_{\mathrm{protect}}$ creates a localized energy barrier while locking the global feature space, establishing a precise boundary that converts target samples into OOD states without disrupting model integrity.


\begin{table}[ht]
\centering
\small

\begin{tabular*}{\columnwidth}{@{\extracolsep{\fill}} lcc @{}} 
\toprule
Metric & Params (M) &  Time (min)\\ 
\midrule
Retrain & 21.60 & 180\\
\textbf{TFER (Ours)} & \textcolor{darkblue}{\textbf{0.24} {\textbf{($\downarrow$ 98.9\%)}}}    & \textcolor{darkblue}{\textbf{9} {\textbf{($\downarrow$ 95.0\%)}}} \\
\bottomrule
\end{tabular*}
\caption{Efficiency comparison between Retrain and TFER.}
\label{tab:efficiency}
\end{table}

\begin{table}[ht]
\centering
\resizebox{\columnwidth}{!}{%
\begin{tabular}{lccc}
\toprule
Loss & Forget FPR95 $\downarrow$ & Retain-Acc $\uparrow$ &  OOD FPR95 $\downarrow$ \\
\midrule

w/o $\mathcal{L}_{\mathrm{protect}}$ & 54.5 & 74.8 & 40.9\\
w/o $\mathcal{L}_{\mathrm{unlearn}}$ & 84.5 & 74.2 & 48.4 \\

\textbf{TFER(Ours)} & \textcolor{darkblue}{\textbf{39.2}} &\textcolor{darkblue} {\textbf{74.9}} & \textcolor{darkblue}{\textbf{40.0}} \\ 

\bottomrule
\end{tabular}
}
\caption{Ablation on loss. All results are percentages (\%).}
\label{tab:ablation_2}
\end{table}

\subsubsection{LoRA Application Location}
\label{sec:ablation_lora}

We compared the performance when applying LoRA to different combinations of layers, as shown in Tab.\ref{tab:ablation_1}. The results clearly indicate that applying LoRA to both Layer4 + Projector (our default setting) achieves the highest overall performance, successfully striking a balance between unlearning efficacy and utility preservation. Specifically, fine-tuning Layer4 alone fails to effectively achieve machine unlearning, resulting in a Forget FPR95 as high as 95.7\%. This suggests that merely altering high-level semantic features without adjusting the downstream mapping is insufficient for reconstructing the decision boundary. Conversely, fine-tuning the Projector alone improves unlearning to some extent, reducing FPR95 to 54.6\%, but comes at the cost of retain set performance, with Retain-Acc dropping to 72.8\%. This indicates that forcing the projection mapping to adapt to fixed feature inputs undermines the model's discriminability on normal samples. These results confirm that joint optimization of the feature extractor and projection head is crucial for maintaining the structural integrity of retained knowledge while stripping away forget samples.

\begin{table}[ht]
\centering

\resizebox{\columnwidth}{!}{%
\begin{tabular}{lccc}
\toprule
LoRA Location & Forget FPR95 $\downarrow$ & Retain-Acc $\uparrow$ &  OOD FPR95 $\downarrow$ \\
\midrule

Layer4 only &95.7 &74.3 &48.4 \\
Projector only & 54.6 & 72.8 &49.9  \\

\textbf{Layer4 + Projector} &\textcolor{darkblue}{ \textbf{39.2}} & \textcolor{darkblue}{\textbf{74.9}} & \textcolor{darkblue}{\textbf{40.0}} \\ 

\bottomrule
\end{tabular}
}
\caption{Ablation on LoRA application location (CIFAR-100). All results are percentages (\%).}
\label{tab:ablation_1}
\end{table}











\subsubsection{Sensitivity to $\lambda_{\mathrm{unlearn}}$}

\begin{figure}[!ht]
    \centering \includegraphics[width=0.48\textwidth]{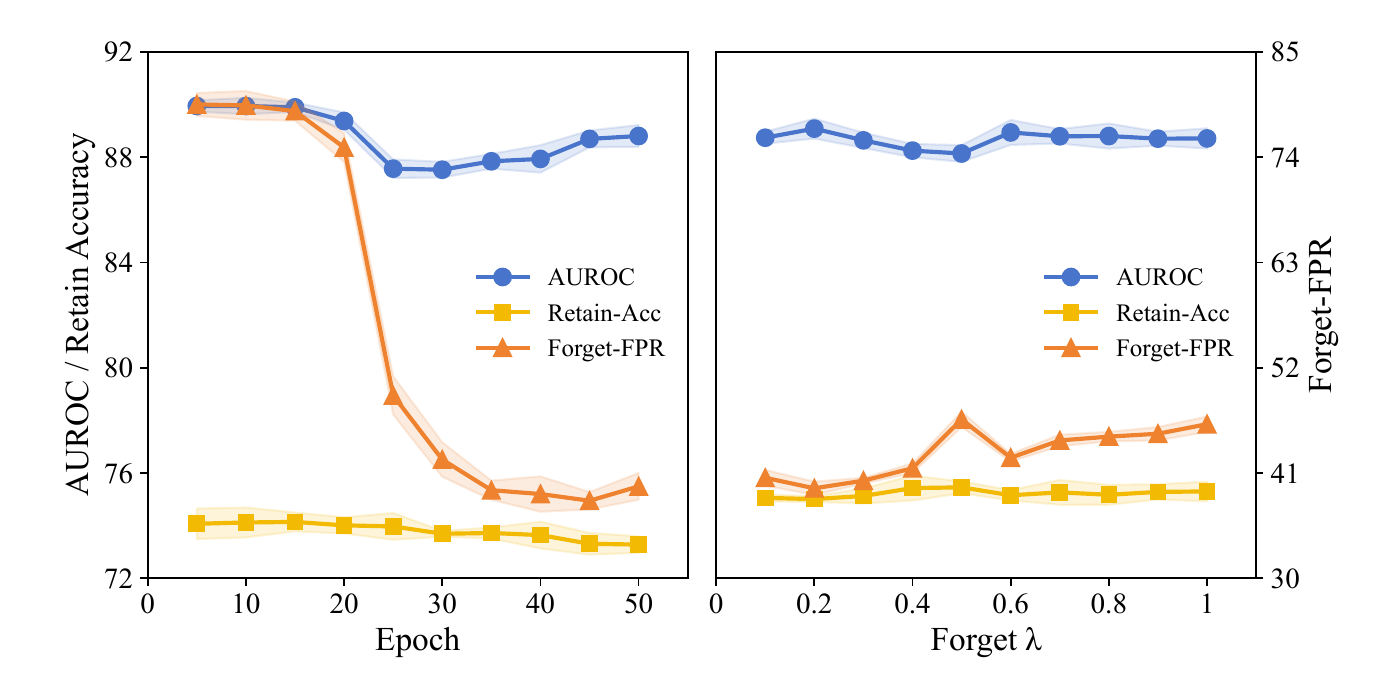} 
        \caption{Sensitivity analysis of TFER performance across different hyperparameters.(a) Impact of training epochs.(b) Impact of forget coefficient.
}
    \label{fig:epoch_rank}
\end{figure}

As illustrated in Fig.\ref{fig:epoch_rank}(b), the model performance remains remarkably stable across a wide range of $\lambda_{\mathrm{unlearn}}$ values. This phenomenon highlights the superiority of the mechanism based on gradient direction dominance and structural decoupling. It reveals that the essence of unlearning lies in the directionality of the optimization vector, specifically, the projection path towards the OOD subspace explicitly defined by the loss function, rather than the gradient magnitude regulated by $\lambda_{\mathrm{unlearn}}$. Once the gradient field is correctly aligned to traverse the decision boundary, increasing the scalar weight merely accelerates saturation without altering the final topological structure. Consequently, we conclude that the dominance of gradient direction prevails over gradient magnitude in effective unlearning.

\subsection{Exploration of Continual Unlearning.}



To simulate real-world scenarios where unlearning requests arrive sequentially, we design a dual-task experiment. Let Task 1 involve the forgetting set $\mathcal{Y}_{f1}$ and Task 2 involve $\mathcal{Y}_{f2}$, where $\mathcal{Y}_{f1} \cap \mathcal{Y}_{f2} = \emptyset$. We compare two strategies: \textbf{Naive Continual}, where the model sequentially updates the same LoRA module inherited from Task 1 on Task 2 data; and \textbf{TFER + Orthogonal (Ours)}, where we freeze $\text{LoRA}_1$, introduce a new module $\text{LoRA}_2$, and enforce an orthogonality constraint $\mathcal{L}_{\text{Orth}}$ between their weight matrices to minimize subspace interference.

As presented in Tab.\ref{tab:continual}, the Naive Continual strategy suffers from severe parameter interference and Catastrophic Recall. Specifically, optimizing for Task 2 not only yields suboptimal performance on the current target ($F_2$ FPR of 73.2) but also causes the unlearning efficacy for the historical target $\mathcal{Y}_{f1}$ to collapse, with the $F_1$ FPR rebounding to near-original levels. In stark contrast, our Modular Orthogonal strategy effectively mitigates this trade-off via parametric isolation. It successfully unlearns the current target $\mathcal{Y}_{f2}$ while robustly preserving the unlearning effects on the historical target $\mathcal{Y}_{f1}$. This validates the scalability of TFER's modular design in handling sustainable, sequential unlearning tasks.








\begin{table}[ht]
\centering

\resizebox{\columnwidth}{!}
{%
\begin{tabular}{lccc}
\toprule
Strategy & Retain-Acc $\uparrow$ & Forget $F_2$ FPR95 $\downarrow$ & Forget $F_1$ FPR95 $\downarrow$ \\
& ($\mathcal{Y}_{\text{Retain}}$) & (Current Task) & (Historical Task) \\
\midrule
Random Label & 73.1 & 46.0 & 46.6 \\
GradAsc      & 74.2 & 68.7 & 62.2 \\

Naive Continual & 73.3 & 73.2 &84.0\\


Original & 74.3 & 84.2 & - \\

Retrain & 41.8 & 95.4 & - \\


\rowcolor{gray2}\textbf{TFER + Orthogonal} & \textcolor{darkblue}{\textbf{74.1}} & \textcolor{darkblue}{\textbf{41.2}} & \textcolor{darkblue}{\textbf{42.1}} \\

\bottomrule
\end{tabular}
}
\caption{Continual unlearning performance (CIFAR-100, 10+10 classes). For static baselines (Original/Retrain), $F_1$ performance is omitted as it is identical to $F_2$ due to the lack of sequential updates. All results are percentages (\%).}
\label{tab:continual}
\end{table}

\section{Conclusion}
In this paper, we address the critical yet previously overlooked problem of boundary-preserving class unlearning in OOD detection models. We identify a fundamental geometric contradiction: traditional classification-centric unlearning methods indiscriminately disrupt the delicate feature manifold, leading to the collapse of the discriminative boundary between ID and anomalous samples. To resolve this dilemma, we introduce the TFER framework, which shifts the unlearning paradigm from simple accuracy degradation to a novel Forget-as-OOD perspective.

TFER implements a Push-Pull dynamic mechanism via parameter-efficient LoRA modules, effectively expelling target classes into high-energy OOD regions while anchoring the retained manifold through a protective pull mechanism. Furthermore, our exploration into continual unlearning demonstrates that TFER’s orthogonal modular design enforces parametric isolation, enabling the model to effectively prevent catastrophic recall.

\bibliographystyle{named}
\bibliography{ijcai26}

\end{document}